\pdfoutput=1
\documentclass[11pt]{article}

\usepackage[final]{acl} %review

\usepackage{times}
\usepackage{latexsym}

\usepackage[T1]{fontenc}
\usepackage[utf8]{inputenc}

\usepackage{microtype}

\usepackage{inconsolata}
\usepackage{adjustbox}

\usepackage{graphicx}
\usepackage{amsmath}
\usepackage{array}
\usepackage{multirow}
\usepackage{arydshln} 
\usepackage{tabularx}

\title{Enhancing Multilingual Embeddings via Multi-Way Parallel Text Alignment}

\author{
 \textbf{Barah Fazili\textsuperscript{1*}\thanks{Research conducted during an internship at Adobe Research in 2024.}},
 \textbf{Koustava Goswami\textsuperscript{2}},
\\
\\
 \textsuperscript{1}IIT Bombay, Mumbai, India
 \textsuperscript{2}Adobe Research, Bangalore, India
\\
}

\begin{document}
\maketitle
% \renewcommand{\thefootnote}{} % Remove footnote numbering
% \footnotetext{Authors marked with * were interns at Adobe Research when this work was started.}
\begin{abstract}
Multilingual pretraining typically lacks explicit alignment signals, leading to suboptimal cross-lingual alignment in the representation space. In this work, we show that training standard pretrained models for cross-lingual alignment with a \emph{multi-way parallel} corpus in a diverse pool of languages can substantially improve multilingual and cross-lingual representations for NLU tasks. We construct a multi-way parallel dataset using translations of English text from an off-the-shelf NMT model for a pool of six target languages and achieve strong cross-lingual alignment through contrastive learning. This leads to substantial performance gains across both seen and unseen languages for multiple tasks from the MTEB benchmark--evaluated for XLM-Roberta and multilingual BERT base models. Using a multi-way parallel corpus for contrastive training yields substantial gains on bitext mining (+21.3\%), semantic similarity (+5.3\%), and classification (+28.4\%) compared to English-centric (En–X) bilingually parallel data, where X is sampled from a pool of multiple target languages. Furthermore, finetuning mE5 model on a small dataset with multi-way parallelism significantly improves bitext mining compared to one without, underscoring the importance of multi-way cross-lingual supervision even for models already pretrained for high-quality sentence embeddings.

\end{abstract}
\section{Introduction}
%what all has been tried to improve the multilingual/cross-lingual performance?
%then segue into CLA (refer to survey paper)
% "Classification tasks are not
% the main strength of generative models, and finetuned encoder models often do better there (Lin
% et al., 2022). \cite{lin-etal-2022-shot}"

Pretrained multilingual language models \cite{conneau-etal-2020-unsupervised} have demonstrated strong transfer capabilities both across tasks in zero-shot settings (e.g bitext mining) and tasks involving finetuning on source data (primarily in English) (e.g. classification). Among the various factors studied behind this emergent behavior—such as shared tokens across some languages—cross-lingual alignment has been shown to be strongly correlated with downstream task performance in cross-lingual tasks~\cite{deshpande-etal-2022-bert,tang2022alignmlmwordembeddingalignment}. 

Cross-lingual alignment implies that "similar meanings across languages have more similar representations than dissimilar meanings do"~\cite{hammerl-etal-2024-understanding}. In the context of natural language understanding tasks, this principle suggests that if a model can effectively map semantically similar expressions from different languages closer in the representation space, then supervision in one language can generalize to others, enabling cross-lingual transfer. There have been numerous attempts at achieving better alignment, (see Section \ref{sec:related-work}) by leveraging parallel text or bilingual dictionaries, but mostly the focus has been on \emph{bilingually} parallel text, even when more languages are included in the corpus. We aim to harness the richer signal of \emph{multi-way parallelism} by introducing \emph{multi-way cross-lingual alignment}, using translations generated by an off-the-shelf machine translation model. We summarize our contributions below:
\begin{enumerate}
    \item We show that using a \emph{multi-way parallel} corpus spanning a pool of target languages can be more effective for cross-lingual alignment training than relying solely on bilingually parallel text.
    \item We treat each language as an anchor during contrastive learning, ensuring that every language actively pulls its translations closer in the embedding space. This approach is a key factor in the success of our \emph{multi-way cross-lingual alignment} training.
    \item We show that training incorporating more languages beside a target language for alignment (wrt English) can lead to better downstream performance across multiple tasks with Hindi evaluated as the target language. 
    \item We also observe that post alignment, our model representations lead to significant gains even for languages unseen during alignment training. This suggests that with our proposed multi-way parallel synthetic corpus, cross-lingual alignment benefits can generalize beyond the seen languages.
    \item The evaluation is done over multiple tasks from MTEB ~\cite{muennighoff2023mtebmassivetextembedding} benchmark for sentence embeddings. Using XLM-R base as the model, we show that leveraging multi-way parallelism in the alignment dataset yields substantial gains across diverse tasks: 21.3\% improvement in bitext mining, 5.3\% in semantic textual similarity, and 28.4\% in classification.
 \end{enumerate}
% In this work, we investigate a few aspects of cross-lingual alignment training and propose the same contrastive alignment objective but over multi-way parallel and synthetic text followed by evaluation on various tasks in MTEB benchmark for sentence embeddings.
% \\what is CLA? strong vs weak
% \\highlight parallelism and more aligned languages, more languages as anchor
% \\Highlight we don't align only wrt English although we maintain English in each row.
% \\highlight that the improvement is significant for a relatively much smaller amount of parallel corpus, labse etc use millions of sentences?
% \\put stress on the fact that we should bring everything closer in a more flexible way rather than pushing all languages towards English. Is there a mathematical support for this?
\section{Methodology}

%\subsection{Multilingual Alignment Dataset and Training}
%Talk about why multilingual alignment can help the multilingual/cross-lingual attribution task.
In the context of images,~\citet{khosla2021supervisedcontrastivelearning} proposed the \textit{supervised contrastive loss}, which extends the \textit{self-supervised} approach by incorporating class information. Instead of relying solely on augmented versions of an anchor image, this approach also considers other instances from the same class within a batch as positive examples, while treating all remaining instances as negatives.

We adapt this idea of allowing multiple positives during contrastive learning to the domain of multilingual text representation learning. Consider a multi-way parallel text corpus \(\mathcal{D}\) with \(k\) fields, with row \(\mathcal{D}_i = \{D_{i,j}\} \; \forall j \in \{1, \dots, k\}\) containing source text in English and its translations in multiple languages over the following $k-1$ fields. For text in each column (in a distinct language across the row) as an anchor, the remaining \(k-1\) columns serve as positives. Here, the notion of \emph{class} is based on semantics, with the translation acting as the augmentation.

\begin{figure}[h]
  \centering
  \includegraphics[width=\linewidth, trim=80 200 200 200, clip]{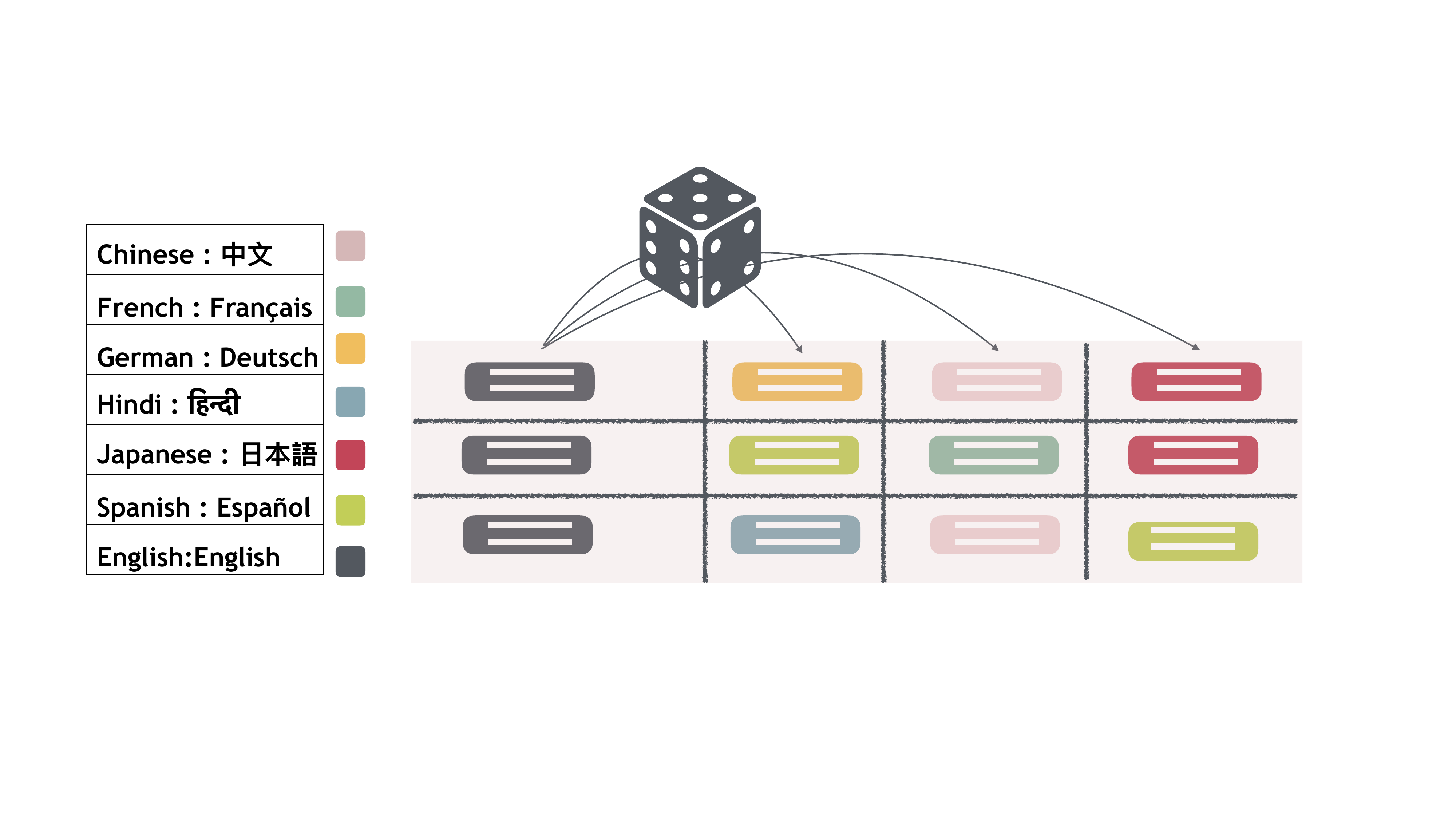}
  \caption{Each sentence (shown as a black cell) in English is translated into three target languages chosen randomly from the pool.}
  \label{fig:dataset-construction}
\end{figure}

We created this corpus of multi-way parallel text after sampling English text from OPUS which is then machine translated.
The created corpus contains four columns and approximately ~75K rows (75,822 in train+validation set). The first column consists of original sentences from OPUS in English, while the other three columns contain translations into three of six target languages—Chinese, Japanese, French, German, Hindi, and Spanish—sampled uniformly at random (see Figure \ref{fig:dataset-construction} ).

Equation \ref{eq:sup_loss} shows the formulation of loss with $P(i)$ indicating the positives of a sentence $i$ with embedding $z_i$ and $A(i) \equiv I \setminus \left( \{i\} \cup P(i) \right)$ the set of remaining sentences in the batch excluding $i$ and the translations of $i$ and $\tau$ is the temperature\footnote{Note that we enforce \emph{strong} alignment~\cite{hammerl-etal-2024-understanding}, ensuring that a sentence’s nearest neighbors in the embedding space are its translations, while (semantically) dissimilar  sentences in the \emph{same} and different languages are spaced farther apart.}. \textit{Instead of using only the English sentence as the anchor in contrastive loss, we allow all languages to serve as the anchor for each row(semantic instance) in our alignment dataset. This approach improves convergence across language subspaces and leads to more cohesive multilingual representations} 

% \begin{align}
%     \mathcal{L}_{sup} &= \sum_{i \in I} \mathcal{L}_i \notag \\
%     &= \sum_{i \in I} \frac{-1}{|P(i)|} \sum_{p \in P(i)} 
%     \log \frac{\exp(z_i \cdot z_p / \tau)}{\sum_{a \in A(i)} \exp(z_i \cdot z_a / \tau)}
% \label{eq:sup_loss}
% \end{align}
\begin{align}
    \mathcal{L}_{sup} &= \sum_{i \in I} \mathcal{L}_i \notag \\
    &= \sum_{i \in I} \frac{-1}{|P(i)|} \sum_{p \in P(i)} 
    \log \frac{\exp(z_i \cdot z_p / \tau)}
    {\sum\limits_{a \in A(i)} \exp(z_i \cdot z_a / \tau)}
\label{eq:sup_loss}
\end{align}

\section{Experiments}
\subsection{Dataset Creation}
\subsubsection{Sampling English Sentences}  
\label{curation}
To construct a diverse set of English source sentences representative of both formal and conversational domains, we combined data from Wikipedia and OpenSubtitles. Specifically, we obtained formal text from Wikipedia by downloading the English corpus from the OPUS-Wikipedia dataset\footnote{\url{https://object.pouta.csc.fi/OPUS-Wikipedia/v1.0/mono/en.txt.gz}}, originally containing over 441M sentences. For conversational text, we used OpenSubtitles, acquiring the English monolingual data from OPUS-OpenSubtitles v2018\footnote{\url{https://opus.nlpl.eu/OpenSubtitles/de&en/v2018/OpenSubtitles}} \footnote{\url{https://object.pouta.csc.fi/OPUS-OpenSubtitles/v2018/mono/en.txt.gz}}, which contained over 10M sentences. To ensure a balanced representation of both styles, we randomly sampled 38,000 sentences from each dataset (with a minimum length of 10 words), using a fixed random seed (42). The sampled subsets were then concatenated into a single corpus. We allocated 90\% of the sentences for training and the remaining 10\% for validation. 
\subsubsection{Translation}  
The selected English sentences were then translated into a diverse set of six target languages—French (\texttt{fra\_Latn}), German (\texttt{deu\_Latn}), Spanish (\texttt{spa\_Latn}), Japanese (\texttt{jpn\_Jpan}), Chinese (\texttt{zho\_Hans}), and Hindi (\texttt{hin\_Deva})—using the NLLB-200 3.3B translation model\footnote{\url{https://huggingface.co/facebook/nllb-200-3.3B}}. The text for each language was maintained in its original script. 
Our setup does not rely on human-curated multi-way corpora, which are rare. Instead, our approach can be easily replicated in any setting where reasonable English-to-target language NMT models exist—a condition increasingly satisfied across many languages due to advances in neural MT. Moreover, while the data is English-rooted, our alignment strategy does not rely on English as a fixed pivot. 
% We explicitly show that allowing all languages to act as anchors (instead of restricting to English) leads to superior alignment and better downstream performance. This ensures that the learned representations are not overly English-centric.

% English, French, Spanish, German, and Hindi all originate from the Indo-European family, representing distinct subgroups: French and Spanish are Romance languages, English and German are Germanic, and Hindi is part of the Indo-Aryan branch. In contrast, Chinese and Japanese come from different families—Sino-Tibetan and Japonic, respectively—but their long history of cultural exchange has resulted in shared vocabulary and some stylistic and syntactic similarities. 

\subsection{Model Training}
We train XLM-Roberta and mBERT-base models\footnote{1. \url{https://huggingface.co/FacebookAI/xlm-roberta-base} 2. \url{https://huggingface.co/google-bert/bert-base-multilingual-cased}} using a contrastive objective over a multi-way parallel corpus. A regularization term (see Equation~\ref{equation:regularizer}) is added to the contrastive loss (Equation~\ref{eq:sup_loss}), which penalizes the squared Euclidean distance between the model embeddings and the corresponding pretrained embeddings ($z_{\text{orig},i}$), normalized over all sentences in the batch ($N = \sum_{i \in I}{1}$). The strength of this regularization is controlled by a scalar hyperparameter $\lambda$.

% While recent MTEB leaderboard entries are dominated by large decoder-style or hybrid models, encoder-only models continue to be strong contenders for multilingual NLU, particularly in resource-constrained environments. 
As shown by \citet{saattrup-nielsen-etal-2025-encoder}, encoder models can outperform decoder models on NLU tasks despite having significantly fewer parameters. %This advantage holds across languages and is not correlated with the resource availability of a language, making encoder models broadly applicable.
Our work is situated in this context, offering a lightweight method for improving encoder models’ cross-lingual transfer performance in real-world applications where inference cost, latency, and model size matter (e.g., edge deployment, multilingual assistants in low-connectivity regions). Additional training details are provided in Appendix~\ref{app:training_details}.

\begin{equation}
\mathcal{R} = \frac{\sum_i \|{z}_{orig,i} - {z}_i\|_2^2}{N} 
\label{equation:regularizer}
\end{equation}
\begin{equation}
\mathcal{L} = \mathcal{L}_{sup} +\lambda\mathcal{R}
\label{equation:full_loss}
\end{equation}
%(using a synthetic data set for attribution task created using Verifiability dataset)
\subsection{Evaluation Results}.

To assess the quality of learned embeddings for multilingual tasks, we evaluate over four different tasks from the Massive Text Embedding Benchmark (MTEB)~\cite{muennighoff2023mtebmassivetextembedding}. We report results for languages included in the dataset used for alignment training, namely English (\textit{en}), French (\textit{fr}), German (\textit{de}), Spanish (\textit{es}), Hindi (\textit{hi}), Chinese (\textit{zh}), and Japanese (\textit{jp}). While bitext mining and STS are evaluated in a zero-shot manner, classification and clustering involve additional task-specific training using MTEB’s default settings, with scores presented on the respective test sets unless stated otherwise.  
\footnote{Datasets can be found here:\url{https://huggingface.co/mteb} and evaluation scripts on the MTEB repo:\url{https://github.com/embeddings-benchmark/mteb/tree/main}. See \url{https://github.com/embeddings-benchmark/mteb/blob/main/docs/tasks.md} for language abbreviation details.}

\subsubsection{Bitext mining}
The input consists of two sets of sentences, each from a different language. The goal is to identify the best match in the second set for each sentence in the first set, typically corresponding to translations. A given model is used to generate sentence embeddings, and the closest pairs are determined based on cosine similarity. Prior work \cite{hu2020xtrememassivelymultilingualmultitask,feng2022languageagnosticbertsentenceembedding} highlights the weak performance of pretrained multilingual models on bitext retrieval tasks due to the absence of a sentence-level objective. As shown in Table \ref{table:bm}, our aligned models (mBERT-aligned and XLM-R-aligned) significantly outperform their standard pretrained counterparts (mBERT-base and XLM-R base) on both BUCC and Tatoeba. With cross-lingual alignment using just 450K sentence pairs, F1 scores improve dramatically (e.g. from 21.6 to 95 for Chinese-English in BUCC for XLMR) across test sets spanning subsets of the pool of target languages.
\subsubsection{Semantic textual similarity (STS)}
The goal is to assess the similarity of a given sentence pair, where labels represent continuous scores, with higher values indicating greater similarity. The provided model generates embeddings for the sentences, and their similarity is measured using cosine values. These computed values are evaluated against ground truth similarity scores using Spearman correlation. Table \ref{table:sts} shows the scores for STS17 and STS22.v2 from the MTEB benchmark. The evaluation sets include both same language(e.g. en-en, es-es) and different languages in the two sets (e.g es-en) in Table \ref{table:sts}. We again see dramatic improvements for this task with the Spearman correlation coefficient going from as low as -1.2 to 52.7 for en-de after alignment of pretrained mBERT-base. Both monolingual and cross-lingual test sets for both datasets get vast improvement across the evaluated languages.
% \textcolor{red}{Table 12 in MTEB paper seems to show *100 of positives values, negative SC values reported as such}
\subsubsection{Classification}
Unlike bitext mining and STS, classification task requires training a classifier head with the sentence embeddings from the model fed as input. A logistic regression classifier, trained on the embeddings from the training set with a maximum of 100 iterations, is evaluated on the test set. Amazon Counterfactual consists of customer review sentences labeled for counterfactual detection (CFD) as a binary classification task. AmazonReviews categorizes reviews into five classes corresponding to the available rating options. MassiveIntent and MassiveScenario contain 60 and 51 labels, respectively, while MTOP Domain has 11 classes, and MTOP Intent spans a variable range of 0 to 112 labels. Table \ref{table:classification} shows the accuracy across six different datasets for languages in our alignment pool where we also indicate the steep percentage gain of finetuned XLMR-aligned model with respect to its counterpart without alignment (finetuned XLMR-base).
\subsubsection{Clustering}
A mini-batch \textit{k}-means model with a batch size of 32 and \textit{k} set to the number of unique labels is trained on the embedded text representations~\cite{pedregosa2011scikit} . The model's performance is evaluated using \textit{V-measure} \cite{rosenberg-hirschberg-2007-v}. Table \ref{table:clustering} shows the scores for five datasets.
With the exception of model initialized to XLMR-aligned performing worse than the pretrained version for SpanishNewsP2P, both aligned models generally perform significantly better across tasks and languages.

% \textcolor{red}{For SpanishNewsP2P, aligned XLMR does worse than the standard pretrained.}

% Note that we don't compare against the SOTA models for any task here since our objective is only to study the proposed alignment technique over the pretrained versions of the respective models with using a much smaller (multi-way) parallel corpus as compared to scale of data typically used for better performing models on the leaderboards e.g  LASER, LABSE etc which use 223 million parallel sentences and 6 billion translation pairs respectively.
\begin{table*}[!htbp]
\centering
\begin{adjustbox}{max width=\textwidth}
\begin{tabular}{|p{0.2\linewidth}|>{\centering\arraybackslash}p{0.05\linewidth}| >{\centering\arraybackslash}p{0.05\linewidth}|>{\centering\arraybackslash}p{0.05\linewidth}|>{\centering\arraybackslash}p{0.05\linewidth}|>{\centering\arraybackslash}p{0.05\linewidth}|>{\centering\arraybackslash}p{0.05\linewidth}|>{\centering\arraybackslash}p{0.05\linewidth}|>{\centering\arraybackslash}p{0.05\linewidth}|>{\centering\arraybackslash}p{0.05\linewidth}|}
\hline 
\multirow{2}{*}{\small{\textbf{Model}}} & \multicolumn{3}{|c|}{\small{\textbf{BUCC}}} & \multicolumn{5}{|c|}{\small{\textbf{Tatoeba}}} \\
\cline{2-9} 
& \small{\textbf{fr-en}} & \small{\textbf{de-en}} & \small{\textbf{zh-en}} & \small{\textbf{de-en}} & \small{\textbf{jp-en}} & \small{\textbf{es-en}} & \small{\textbf{fr-en}} & \small{\textbf{hi-en}} \\
\hline 
\small{mBERT-base} & \small{10.8} &\small{11.3} &\small{4.9} & \small{9.4}& \small{6.9}&\small{11.5}&\small{14.6} &\small{2.3}\\
\small{mBERT-aligned} &\small{91.4} &\small{95} &\small{91.7}& \small{91.8}& \small{76.2}&\small{84.5}&\small{83.2} &\small{73.6} \\
\small{XLMR-base}&\small{18.2} &\small{26.9} &\small{21.6} & \small{32}& \small{13.5}&\small{23.8} &\small{22.4} &\small{9.9} \\
\small{XLMR-aligned} &\small{92.9}&\small{95.9} &\small{95}& \small{95.7}& \small{86.3}&\small{92.5}&\small{89.6} &\small{89.4} \\
\hdashline
\small{XLMR-en-anchor} &\small{23.2}&\small{37.1} &\small{40.1}& \small{47.8}& \small{29.4}&\small{48.9}&\small{37.8} &\small{28.8} \\
\small{XLMR-en-ablate} &\small{87.9}&\small{93.4} &\small{93.3}& \small{90}& \small{72.9}&\small{81.1}&\small{77.9} &\small{68.1} \\
\hline 
\hline 
\small{par-model-A} & \small{94.8} &\small{65.7} &\small{96.8} & \small{96}& \small{87.8}&\small{92.7}&\small{91.2} &\small{90}\\
\small{par-model-B} &\small{80.7} &\small{83.4} &\small{86.8}& \small{83.6}& \small{63.8} & \small{72.6}&\small{71.4}&\small{58.1} \\
\hline
\end{tabular}
\end{adjustbox}
\caption{\label{table:bm}Results for bitext mining}
\end{table*}

\begin{table*}[t!]
\centering
\begin{adjustbox}{max width=\textwidth}
\begin{tabular}{|p{0.18\linewidth}|>{\centering\arraybackslash}p{0.05\linewidth}|>{\centering\arraybackslash}p{0.05\linewidth}|>{\centering\arraybackslash}>{\centering\arraybackslash}p{0.05\linewidth}|>{\centering\arraybackslash}p{0.05\linewidth}|>{\centering\arraybackslash}p{0.05\linewidth}|>{\centering\arraybackslash}p{0.05\linewidth}|>{\centering\arraybackslash}p{0.05\linewidth}|>{\centering\arraybackslash}p{0.05\linewidth}|>{\centering\arraybackslash}p{0.05\linewidth}|>{\centering\arraybackslash}p{0.05\linewidth}|>{\centering\arraybackslash}p{0.05\linewidth}|>{\centering\arraybackslash}p{0.05\linewidth}|>{\centering\arraybackslash}p{0.05\linewidth}|>{\centering\arraybackslash}p{0.05\linewidth}|}
\hline 
\multirow{2}{*}{{\textbf{Model}}}& \multicolumn{5}{|c|}{{\textbf{STS17}}} & \multicolumn{9}{|c|}{{\textbf{STS22.v2}}} \\
\cline{2-15} 
& {\textbf{en-en}} & {\textbf{fr-en}} & {\textbf{en-de}} & {\textbf{es-es}} & {\textbf{es-en}} & {\textbf{en}} & {\textbf{fr}} & {\textbf{de}} & {\textbf{es}} & {\textbf{zh}} & {\textbf{de-en}} & {\textbf{de-fr}} & {\textbf{es-en}} & {\textbf{zh-en}}\\
\hline 
{mBERT-base} &{24.9} &{2.9}&{$-$1.2} &{29.2} & {6.8}& {3.2} &{10} &{$-$24.5} &{1} & {$-$1.7}& {$-$4.7}&{4.7} &{2.5} &{3.9}\\
{mBERT-aligned} &{65.3} &{57.1}&{52.7} &{69.5} & {43.6}& {54.9} &{67.9} &{37} &{60.2} & {65.1}& {51.3}&{43.5} &{68.6} &{62.3}\\
{XLMR-base} &{19.9} &{2.79}&{4.8} &{20.5} & {4.4}& {50.8} &{62.7} &{19.7} &{50.9} & {54.9}& {42.8}&{38.9} &{40.3} &{45.2}\\
{XLMR-aligned} &{63.7} &{55.2}&{51.9} &{70.6} & {47.6}& {61.7} &{75.2} &{41.7} &{66.8} & {68.4}& {56.9}&{49.6} &{71.3} &{66.6}\\
\hdashline
{XLMR-en-anchor} &{35.2} &{9.4}&{21.6} &{32.5} & {31}& {58.5} &{60.9} &{30.2} &{60.6} & {67.2}& {57.3}&{48.8} &{63.4} &{60.7}\\
{XLMR-en-ablate} &{70.1} &{55.2}&{57} &{74.8} & {58.7}& {57.8} &{67} &{33.7} &{64.6} & {68.3}& {51.1}&{51.9} &{69.1} &{60.9}\\
\hline
\hline 
{par-model-A} &{66.9} &{58.4}&{58.7} &{73.9} & {52.5}& {58.2} &{73} &{43.6} &{65.7} & {67.7}& {52.8}&{54.2} &{70} &{63.4}\\
{par-model-B} &{65.6} &{47.8}&{48.7} &{68.8} & {45.4}& {58.4} &{68.6} &{41.4} &{66} & {68.5}& {54.7}&{52} &{70.9} &{64.9}\\
\hline 
\end{tabular}
\end{adjustbox}
\caption{\label{table:sts}Results for STS (semantic textual similarity)}
\end{table*}

% \begin{table*}[t!]
\begin{table*}[ht]
\centering
\begin{adjustbox}{max width=\linewidth}
\begin{tabular}{|p{0.25\linewidth}|>{\centering\arraybackslash}p{0.08\linewidth}| p{0.08\linewidth}|>{\centering\arraybackslash}p{0.08\linewidth}|>{\centering\arraybackslash}p{0.08\linewidth}|>{\centering\arraybackslash}p{0.15\linewidth}|>{\centering\arraybackslash}p{0.08\linewidth}|>{\centering\arraybackslash}p{0.08\linewidth}||>{\centering\arraybackslash}p{0.08\linewidth}|>{\centering\arraybackslash}p{0.08\linewidth}|}
\hline 
\textbf{Dataset} & \textbf{Lang} & \textbf{mBERT-base} & \textbf{mBERT-aligned} & \textbf{XLMR-base} & \textbf{XLMR-aligned (\% gain)} & \textbf{XLMR-en-anchor} & \textbf{XLMR-en-ablate} & \textbf{par-model-A} & \textbf{par-model-B} \\
\hline 
\multirow{3}{*}{\rotatebox{0}{\textbf{AmazonCounterfactual}}} & en & {60.6} & {66.8} & {62.5} & {72.0 (15.2)} & {60} & {61.6} &{72.5} & {58.4}  \\
 & de & {59.4} & {65.2} & {55.4} & {70.5 (27.3)}  & {52.3} &{60.9} & {69.7} & {61.2}  \\
 & jp & {57.1} &{59} & {59}& {72.6(23.05)}  & {56} & {58.8} & {72.9} & {56.2}\\
\hline 
\hline 
\multirow{6}{*}{\rotatebox{0}{\textbf{AmazonReviews}}} & en & {27.5} & {32.3} & {26.1} & {36.8 (40.99)}  & {26} & {28.6} & {36.9} & {28.3}  \\
& de & {24.7} & {32.3} & {28.4} & {39.5 (39.08)}  & {27.5}  & {31.6} &{39.6} & {30.9}  \\
& es & {25.8} &{32.4} & {25.1} &{38.2 (52.19)}  & {25.7}  & {31.6} &{38.4} & {30.3} \\
& fr & {25.4} & {31.6} & {25.6}& {38.5 (50.39)}  & {25.2}  & {31.8} &{38.3} & {30} \\
& jp & {26.7} &{29.8} & {26.9}& {36.9 (37.17)}  & {27.7} & {32.8} &{36.7} & {29.8} \\
& zh & {25.8} & {30.7} & {29.4} & {36.5 (24.15)} & {30.3} & {33} &{36.8} & {29.9} \\
\hline 
\hline 
\multirow{6}{*}{\rotatebox{0}{\textbf{MassiveIntent}}} & de & {28.3} & {41.3} & {19.3} & {53.9 (179)}  & {21.2} & {37.5} &{54.6} & {38.7} \\
& es & {29.9} & {46.7} & {22.8} & {56.4 (147)}  & {22.8}  & {41.1} & {56.4} & {42.2} \\
& en & {37.8} &{51.7} & {29.5}& {60.7 (105)} & {29.4} & {45.4} & {61.5} & {46.2} \\
& fr & {31.7} & {47} & {20.2} & {55.9 (176)}  & {21.5} & {40.2} & {56.3} & {42} \\
& jp & {32.9} &{49.5} & {24.2}& {60 (147)}  & {19.7} & {44.9} & {60.2} & {46.4} \\
& hi & {24.5} & {38.6} & {23.4} & {53.8 (129)}  & {17.7} & {41.1} & {54.5} & {40.9} \\
\hline 
\hline 
\multirow{6}{*}{\rotatebox{0}{\textbf{MassiveScenario}}} & en & {40.1} & {57.4} & {39.8} & {66.8 (67.84)} & {39.9} & {59.8} & {67.6} & {62} \\
& hi & {26.8} & {42.1} & {31} & {58.9 (90)} & {26.3}  & {50.7} & {59.4} & {50.7}  \\
& de & {31.2} &{45.9} & {28.7}& {60.9 (112)}  & {31.9} & {52.9} & {62.3} & {52.3}  \\
& fr & {35.6} & {50.9} & {27.6} & {60.6 (119)}  & {30.6} & {54.3} & {61.6} & {56.1}  \\
& jp & {34.8} &{53.1} & {33.5}& {65.3 (94.9)}  & {27.9} & {61.3} & {66.2} & {60.2}  \\
& es & {32.4} & {52} & {32.3} & {62.5 (93.5)} & {32.5} & {54.6} & {63} & {57.3}  \\
\hline 
\hline 
\multirow{5}{*}{\rotatebox{0}{\textbf{MTOPDomain}}} & en & {55} & {72.1} & {39.6} & {80.3 (102)}  & {35.3} & {79.4} & {81.3} & {74.7}  \\
& de & {49} & {69.8} & {40.2} & {80.5 (100)}  & {28.8} & {77.7} & {81.5} & {72.8}  \\
& es & {51} &{70.8} & {37.2}& {78.1 (109)}  & {33.3} & {76.1} & {79.4} & {71.5}  \\
& fr & {49.4} & {66.5} & {38.7} & {74.9 (93.5)}  & {34.5} & {69.9} & {74.8} & {65.4}  \\
& hi & {44.8} &{59.6} & {34.4}& {76.4 (122)} & {31.7} & {72.3} & {76.6} & {68.4}  \\
\hline 
\hline 
\multirow{5}{*}{\rotatebox{0}{\textbf{MTOPIntent}}} & en & {40.1} & {51.7} & {19.5} & {66 (238)}  & {8.8} & {37.4} & {65} & {38}  \\
& de & {35.8} & {41.3} & {20.5} & {65.3 (218)} & {8.1}  & {35.5} & {64.4} & {38} \\
& es & {35.1} &{46.6} & {21.5}& {62 (188)}  & {8.5} & {39.2} & {61.4} & {40}  \\
& fr & {32.1} & {47} & {15.5} & {58 (274)}  & {9.6} & {37.2} & {56.9} & {34}  \\
& hi & {28.4} &{38.6} & {13.4}& {62.1 (363)} & {6.6}  & {37.8} & {61} & {39.3}  \\
\hline 

\end{tabular}
\end{adjustbox}
\caption{\label{table:classification}Results for classification.}
\end{table*}

\begin{table*}[t!]
\centering
\begin{adjustbox}{max width=\linewidth}
\begin{tabular}{|p{0.18\linewidth}|>{\centering\arraybackslash}p{0.18\linewidth}|>{\centering\arraybackslash}p{0.1\linewidth}|>{\centering\arraybackslash}p{0.1\linewidth}|>{\centering\arraybackslash}p{0.1\linewidth}|>{\centering\arraybackslash}p{0.1\linewidth}|>{\centering\arraybackslash}p{0.15\linewidth}|>{\centering\arraybackslash}p{0.18\linewidth}|}
\hline 
% \multirow{2}{*}{\small{\textbf{Model}}}& \multicolumn{1}{|c|}{\small{\textbf{IndicReviewsP2P}}} & \multicolumn{2}{|c|}{\small{\textbf{MasakhaNewsP2P}}} & \multicolumn{2}{|c|}{\small{\textbf{MasakhaNewsS2S}}} & \multicolumn{1}{|c|}{\small{\textbf{MewsC16Ja}}} & \multicolumn{1}{|c|}{\small{\textbf{SpanishNewsP2P}}}\\
\multirow{2}{*}{{\textbf{Model}}} 
& {\textbf{IndicReviewsP2P}} 
& \multicolumn{2}{c|}{{\textbf{MasakhaNewsP2P}}} 
& \multicolumn{2}{c|}{{\textbf{MasakhaNewsS2S}}} 
& {\textbf{MewsC16Ja}} 
& {\textbf{SpanishNewsP2P}} \\
\cline{2-8} 
& {\textbf{hi}} & {\textbf{en}} & {\textbf{fr}}  & {\textbf{en}}  & {\textbf{fr}}  & {\textbf{jp}}  & {\textbf{es}}   \\
\hline 
{mBERT-base} & {23.2} &{0.6} &{25.2} & {1.5}& {22}&{6.8}&{4.5} \\
{mBERT-aligned} &{35.2} &{47.1} &{56.7}& {32.6}& {42.2}&{44.9}&{26.2}  \\
{XLMR-base} &{32}&{16} &{22.7}& {1.2}& {21}&{22.8}&{38.7}  \\
{XLMR-aligned}&{37.8} &{52.2} &{57.2} & {34.7}& {27.1}&{43.4} &{26} \\
\hdashline
{XLMR-en-anchor}&{35.2} &{29.8} &{48} & {1}& 
{20.9}&{34.9} &{47.5} \\
{XLMR-en-ablate}&{42.2} &{52.6} &{63.8} & {38.6}& {53}&{48} &{42.4} \\
\hline 
% \small{par-model-A} & \small{40.8} &\small{44.8} &\small{66.8} & \small{39.8}& \small{30.6}&\small{43.7}&\small{30.5} \\
% \small{par-model-B} &\small{39.3} &\small{44.7} &\small{59.6}& \small{31.6}& \small{43.9}&\small{47.5}&\small{33.2}  \\
% \hline 
\end{tabular}
\end{adjustbox}
\caption{\label{table:clustering}Results for clustering}
\end{table*}

\begin{table}[t!]
% \centering
\begin{adjustbox}{max width=\linewidth}
\begin{tabular}{|>{\centering\arraybackslash}p{0.3\linewidth}|>{\centering\arraybackslash}p{0.3\linewidth}|>{\centering\arraybackslash}p{0.3\linewidth}|}
\hline 
\small{\textbf{Model}} & {\small{\textbf{Tatoeba (hi-en)}}} & {\small{\textbf{IndicReviewsP2P (hi)}}} \\
\hline
\multicolumn{1}{|l|}{\small{multi-eh}}  &\small{69.2}  &\small{41.3} \\
\multicolumn{1}{|l|}{\small{multi-euro}}&\small{67.9} & \small{43.1} \\
\multicolumn{1}{|l|}{\small{multi-eh-asian}}&\small{67.1} &\small{43} \\
\multicolumn{1}{|l|} {\small{multi-eh-all}} &\small{71.4}& \small{43.8} \\
\hline 
\end{tabular}
\end{adjustbox}
\caption{\label{table:multi-hyp-BM-cluster} Results for bitext mining and clustering after bilingual vs multilingual alignment}
\end{table}

% \begin{table}[t!]
% \centering
% \begin{adjustbox}{max width=\textwidth}
% \begin{tabular}{|p{0.3\linewidth}|p{0.05\linewidth}|}
% \hline 
% \small{\textbf{Model}} & \multicolumn{1}{|c|}{\small{\textbf{IndicReviewsP2P (hi)}}} \\
% \hline 
% \small{multi-eh} &\small{0.413} \\
% \small{multi-eh-euro}&\small{0.431} \\
% \small{multi-eh-asian}&\small{0.430} \\
% \small{multi-eh-all}&\small{0.438} \\
% \hline
% \end{tabular}
% \end{adjustbox}
% \caption{\label{table:multi-hyp-clustering}Results for Clustering after bilingual vs multilingual alignment}
% \end{table}

% \begin{table}[t!]
% \centering
% \begin{adjustbox}{max width=\textwidth}
% \begin{tabular}{|p{0.2\linewidth}|>{\centering\arraybackslash}p{0.1\linewidth}|}
% \hline 
% \small{\textbf{Model}} & \small{\textbf{Tatoeba (hi-en)}} \\
% \hline
% \small{multi-eh}  & 0.692 \\
% \small{multi-euro} & 0.679 \\
% \small{multi-eh-asian} & 0.671 \\
% \small{multi-eh-all} & 0.714 \\
% \hline 
% \end{tabular}
% \end{adjustbox}
% \caption{\label{table:multi-hyp-BM} Results for BM after bilingual vs multilingual alignment}
% \end{table}

\begin{table}[t!]
\centering
\begin{adjustbox}{max width=\textwidth}
\begin{tabular}{|p{0.275\linewidth}|
>{\centering\arraybackslash}p{0.07\linewidth}| 
>{\centering\arraybackslash}p{0.07\linewidth}|
>{\centering\arraybackslash}p{0.08\linewidth}|
>{\centering\arraybackslash}p{0.08\linewidth}|
>{\centering\arraybackslash}p{0.08\linewidth}|}
\hline 
\textbf{\small{Dataset}} & \small{\textbf{Lang}}  & \small{\textbf{multi-eh}} & \small{\textbf{multi-eh-asian}} & \small{\textbf{multi-eh-euro}} & \small{\textbf{multi-eh-all}}\\
\hline 
\multirow{1}{*}{\rotatebox{0}{\small{\textbf{MassiveIntent}}}} & hi & \small{37.3} & \small{39} & \small{39.1}  & \small{39.9} \\
\hline 
\hline 
\multirow{1}{*}{\rotatebox{0}{\small{\textbf{MassiveScenario}}}} & hi  & \small{46.8} & \small{50.4} & \small{48.8}  & \small{50.7} \\
\hline 
\hline 
\multirow{1}{*}{\rotatebox{0}{\small{\textbf{MTOPDomain}}}} & hi  & \small{70.8} & \small{71.8} & \small{71.6}  & \small{74.3} \\
\hline 
\hline 
\multirow{1}{*}{\rotatebox{0}{\small{\textbf{MTOPIntent}}}} & hi  & \small{33.4} & \small{37.7} & \small{38}  & \small{39.5} \\

\hline 
\end{tabular}
\end{adjustbox}
\caption{\label{table:multi-hyp-classification}Results for classification after bilingual vs multilingual alignment}
\end{table}

% To address potential domain overlap between the model’s pretraining alignment corpus taken from OPUS and the evaluation test sets, we also pretrain an alignment model using a curated corpus from an unrelated dataset \cite{liu2023evaluatingverifiabilitygenerativesearch} (details of corpus creation in Appendix) 
\section{Analysis}
\subsection{Multi-way Parallelism}
\label{sec:parallelism}
In order to check whether the choice of alignment training over the multi-way parallel corpus is any better than English-centric (En–X) bilingually parallel data, where X is sampled from a pool of multiple target languages, we train XLMR-base model using the following two alignment corpora:
\begin{enumerate}
    \item The model (par-model-A) is trained on a subset of N/6 of the rows in our original dataset $\mathcal{D}$. 
    \item Another model (par-model-B) is trained on a bilingually parallel slice of $\mathcal{D}$ with all N rows but with randomly chosen second column (language) beside the first column in English for alignment. %Note that we maintain anchors in the contrastive loss from each languages in both settings.
\end{enumerate}
To ensure a fair comparison, we retain only N/6 rows for the multi-way parallel setup so that the total number of parallel sentence pairs remains the same across both settings. As shown in Tables \ref{table:bm}, \ref{table:sts} and \ref{table:classification}, the multi-way aligned model (par-model-A) consistently outperforms its bilingually parallel counterpart (par-model-B) across all three evaluated downstream tasks for most languages and language-pairs \footnote{For classification, par-model-A significantly outperforms par-model-B across 31 dataset-language pairs (paired t-test: t = 10.81, p < 1e-10; Wilcoxon: W = 0.0, p < 1e-6), confirming the statistical significance of the improvement.}.
\subsection{Bilingual vs multilingual}
% tested only for Hi(en-hi alignment)
To evaluate whether aligning additional languages beside the target improves performance, we designed four training datasets for Hindi evaluation:
\begin{enumerate}
    \item Bilingual alignment (eh): A corpus containing only English and Hindi (two columns) with  N rows.
    \item European language augmentation (eh-euro): English and Hindi as fixed columns, with the remaining two columns randomly sampled from French, German, and Spanish (four columns, N rows).
    \item Asian language augmentation (eh-asian): English and Hindi as fixed columns, with the remaining two columns sampled from Chinese and Japanese (four columns, N rows).
    \item Full multilingual alignment (eh-all): English and Hindi as fixed columns, with the other two columns sampled from all remaining languages (four columns, N rows).
\end{enumerate}
As shown in Tables \ref{table:multi-hyp-BM-cluster} and \ref{table:multi-hyp-classification}, aligning across multiple languages not only doesn't hurt but can also enhance performance on a specific target language. For bitext mining, adding only European (eh-euro) or only Asian (eh-asian) languages does not improve over bilingual alignment (eh), but combining both groups (eh-all) leads to a performance boost. For other tasks, including classification, clustering, and STS, all multilingual variants (eh-euro, eh-asian, and eh-all) outperform the bilingual baseline, with eh-all consistently achieving the best results.

\subsection{Multiple languages vs English only as anchor}
A key design choice in our alignment training was allowing all languages in the dataset to serve as anchors, with their respective translations acting as positive pairs. To assess the impact of this decision, we trained an alternative model where only English was used as the anchor. The results show a significant drop in performance when restricting the anchor to English alone. As illustrated in Tables \ref{table:bm}, \ref{table:sts}, \ref{table:classification}, and \ref{table:clustering}, the scores for XLMR-en-anchor decline considerably compared to XLMR-aligned, where all languages (present in the corresponding row) were included as anchors. This shows the advantage of leveraging multilingual anchors for better alignment and transferability across languages.

\subsection{Effect of removing mandatory English}
To evaluate whether it is essential to retain English in every semantic instance, we created an alternative dataset where all four languages per instance were randomly sampled from a pool of seven languages (including English), rather than ensuring the presence of English in each row. We trained XLMR-base on this modified dataset, naming the model XLMR-en-ablate. While XLMR-en-ablate performs worse than XLMR-align (our default setting), it still significantly outperforms XLMR-en-anchor, where only English was used as the anchor. This suggests that allowing all languages to act as anchors during contrastive training plays a far more critical role in alignment than strictly maintaining English in every instance.
\subsection{Effect of alignment on other languages}
Beyond achieving substantial gains on test sets for languages included in our alignment training, we also assess its impact on languages outside our selected pool of seven. We evaluate the STS task using XLM-R base and aligned models across all available test languages. Notably, we observe significant and consistent improvements even in languages like Arabic, Russian, and Turkish (see Table \ref{table:sts-non-pool}). We also evaluate on three different clustering tasks (see Table \ref{table:clustering-non-pool}) over twelve Indian languages in IndicReviewsClusteringP2P and fourteen African languages in each of MasakhaNewsClusteringP2P and MasakhaNewsClusteringS2S. With the exception of two cases (\textit{bd} and \textit{lin} in the first and third task respectively), we again see substantial improvements across the board. This highlights how multi-way cross-lingual alignment training over a small subset of diverse languages can potentially generalize to unseen languages.
\begin{table*}[h!]
\centering
\begin{adjustbox}{max width=\textwidth}
\begin{tabular}{|p{0.15\linewidth}|p{0.05\linewidth}| p{0.05\linewidth}|p{0.05\linewidth}|p{0.05\linewidth}|p{0.05\linewidth}|p{0.05\linewidth}|p{0.05\linewidth}|p{0.05\linewidth}|p{0.05\linewidth}|p{0.05\linewidth}|p{0.05\linewidth}|p{0.05\linewidth}|p{0.05\linewidth}|p{0.05\linewidth}|p{0.05\linewidth}|}
\hline 
\multirow{2}{*}{{\textbf{Model}}} & \multicolumn{6}{|c|}{{\textbf{STS17}}} & \multicolumn{9}{|c|}{{\textbf{STS22.v2}}} \\
\cline{2-16} 
& {\textbf{en-ar}} & {\textbf{ar-ar}} & {\textbf{en-tr}} & {\textbf{it-en}} & {\textbf{nl-en}} & {\textbf{ko-ko}} & {\textbf{ar}} & {\textbf{es-it}} & {\textbf{ru}} & {\textbf{it}} & {\textbf{tr}} & {\textbf{fr-pl}} & {\textbf{pl}} & {\textbf{de-pl}} & {\textbf{pl-en}}\\
\hline 
{XLMR-base} &{\-8.8} &{3.9}&{3.2} &{2.2} & {4}& {32.2} &{48.3} &{42.1} &{46.1} & {53.8}& {30.4}&{16.9} &{27.9} &{8.3}  &{27.6}\\
{XLMR-aligned} &{48.3} &{46}&{52.6} &{53.6} & {59.1}& {56.9} &{60.5} &{69.6} &{57.7} & {69.8}& {62.3}&{39.4} &{41} &{50.8}  &{67.3}\\

\hline 
\end{tabular}
\end{adjustbox}
\caption{\label{table:sts-non-pool}Semantic textual similarity scores for languages unseen (during alignment training)}
\end{table*}

% \begin{table}[t!]
\begin{table}[!htbp]
\centering
\begin{adjustbox}{max width=\textwidth}
\begin{tabular}{|p{0.2\linewidth}|>{\centering\arraybackslash}p{0.1\linewidth}| >{\centering\arraybackslash}p{0.2\linewidth}|>{\centering\arraybackslash}p{0.2\linewidth}|}
\hline 
\textbf{\small{Dataset}} & \small{\textbf{Lang}}  & \small{\textbf{XLMR-base}} & \small{\textbf{XLMR-aligned}} \\
\hline 
% \multirow{2}{*}{\rotatebox{90}{\small{\textbf{IndicReviewsClusteringp2p}}}} & \small{as} & 
\multirow{12}{*}{\parbox[c]{1.5cm}{\centering \rotatebox{90}{\small{\textbf{IndicReviewsClusteringp2p}}}}} & \small{as} &
\small{21.8} & \small{30.4}   \\
 & \small{bd} & \small{21.1} & \small{20.4}  \\
 & \small{bn} & \small{29.9} & \small{36.9}  \\ 
& \small{gu} & \small{27.2} & \small{36.3}  \\ 
& \small{kn} & \small{28.3} & \small{37.3}  \\ 
& \small{ml} & \small{25} & \small{37.2}  \\ 
& \small{mr} & \small{24.1} & \small{36.6}  \\ 
& \small{or} & \small{23.3} & \small{33.2}  \\ 
& \small{pa} & \small{27.1} & \small{34.6}  \\ 
& \small{ta} & \small{26.9} & \small{38.1}  \\ 
& \small{te} & \small{28.7} & \small{39.3}  \\ 
& \small{ur} & \small{30.3} & \small{36.2}  \\ 
\hline 
\multirow{14}{*}{\parbox[c]{1.5cm}{\centering \rotatebox{90}{\small{\textbf{MasakhaNewsClusteringp2p}}}}} & \small{amh} & \small{43.1} & \small{60.4}   \\
& \small{hau} & \small{16.9} & \small{61.7}  \\ 
& \small{ibo} & \small{23.3} & \small{40.8}  \\ 
& \small{lin} & \small{46.2} & \small{64.1}  \\ 
& \small{lug} & \small{42.2} & \small{50.9}  \\ 
& \small{orm} & \small{21.5} & \small{30.8}  \\ 
& \small{pcm} & \small{38.5} & \small{66.1}  \\ 
& \small{run} & \small{46.4} & \small{53.4}  \\ 
& \small{sna} & \small{43.4} & \small{45.3}  \\ 
& \small{som} & \small{29.1} & \small{32.2}  \\ 
& \small{swa} & \small{21.8} & \small{31.8}  \\ 
& \small{tir} & \small{42.1} & \small{60.8}  \\ 
& \small{xho} & \small{20.6} & \small{35.9}  \\ 
& \small{yor} & \small{21.1} & \small{33.1}  \\ 
\hline 
\multirow{14}{*}{\parbox[c]{1.5cm}{\centering \rotatebox{90}{\small{\textbf{MasakhaNewsClusterings2s}}}}} & \small{amh} & \small{40.5} & \small{50.3}   \\
& \small{hau} & \small{50} & \small{24.2}  \\ 
& \small{ibo} & \small{22.4} & \small{30.8}  \\ 
& \small{lin} & \small{45.9} & \small{43.1}  \\ 
& \small{lug} & \small{40.5} & \small{42.3}  \\ 
& \small{orm} & \small{20.4} & \small{20.7}  \\ 
& \small{pcm} & \small{24.1} & \small{52}  \\ 
& \small{run} & \small{41.9} & \small{49.5}  \\ 
& \small{sna} & \small{40.4} & \small{41.7}  \\ 
& \small{som} & \small{22.1} & \small{27}  \\ 
& \small{swa} & \small{4.6} & \small{9}  \\ 
& \small{tir} & \small{41.4} & \small{47.8}  \\ 
& \small{xho} & \small{23.3} & \small{24.9}  \\ 
& \small{yor} & \small{21.2} & \small{32.8}  \\ 
\hline 
\end{tabular}
\end{adjustbox}
\caption{\label{table:clustering-non-pool}Clustering scores for out-of-pool languages }
\end{table}
\subsection{On retrained sentence embedding models:bitext mining}
% It may be argued that out-of-the-box performance on sentence embedding tasks with general-purpose multilingual encoders like mBERT and XLM-R is already known to be poor, and that comparisons should therefore prioritize models like multilingual E5 (ME5), which are explicitly trained for sentence-level semantic tasks. We agree with this reasoning and include ME5 in our evaluation.

% We find that aligning ME5 using our multi-way parallel dataset leads to improved performance on bitext mining tasks, where precise cross-lingual sentence-level alignment is crucial. Our fine-tuned variant of ME5 consistently outperforms the original ME5 model on this task, highlighting the value of high-quality multi-way supervision even for already well-aligned models (See Table \ref{table:bm-me5}).

% This result is particularly notable given that our fine-tuning setup is relatively lightweight—relying on a simple contrastive loss over ~60K examples—compared to ME5’s original multi-stage training pipeline involving massive multilingual corpora, knowledge distillation, and instruction tuning. The improvements we observe suggest that multi-way supervision can serve as a valuable complement to such large-scale training, especially in tasks that explicitly benefit from richer cross-lingual alignment.
We also experiment with mE5~\cite{wang2024multilinguale5textembeddings} as the model. Despite mE5’s strong prior pretraining (through large-scale weakly supervised contrastive pretraining on ~1B multilingual pairs, followed by supervised fine-tuning with hard negatives and knowledge distillation), we find that fine-tuning it with our method over a small multi-way parallel dataset improves performance on bitext mining tasks. Our multi-way aligned mE5 variant (mE5-par-model-A) outperforms both the original model (mE5-base) on this task and alignment over English-centric bilingually parallel dataset (mE5-par-model-B) (see Table \ref{table:bm-me5}), demonstrating that even for well-aligned models, multi-way cross-lingual supervision adds value~\footnote{A paired t-test for mE5-par-model-A vs mE5-par-model-B scores yields a t-statistic of 4.44 with a p-value of 0.003, and Wilcoxon signed-rank test gives W = 0.0 with p-value = 0.0078 indicating a statistically significant difference.}. Given that our fine-tuning uses only $\sim{10K}$ rows from our dataset with supervised contrastive loss, this result underscores the strength of the multi-way signal and its potential as a lightweight, complementary strategy to large-scale training. These gains demonstrate the strength of our alignment strategy, and further improvements are likely with larger corpora and tailored training methods.
% avg_bm_me5_improvement in percentage: 0.715949268629059
% Paired t-test: t-statistic for BM mE5 = 4.4388, p-value = 0.0030
% Wilcoxon signed-rank test: W  for BM mE5 = 0.0, p-value = 0.0078
% Further details, including results on classification are included in Appendix \ref{app:me5}.
% \begin{table}[!htbp]
% \centering
% \begin{adjustbox}{max width=\textwidth}
% \begin{tabular}{|p{0.18\linewidth}|p{0.05\linewidth}| p{0.05\linewidth}|p{0.05\linewidth}|p{0.05\linewidth}|p{0.05\linewidth}|p{0.05\linewidth}|p{0.05\linewidth}|p{0.05\linewidth}|p{0.05\linewidth}|}
% \hline 
% \multirow{2}{*}{\small{\textbf{Model}}} & \multicolumn{3}{|c|}{\small{\textbf{BUCC}}} & \multicolumn{5}{|c|}{\small{\textbf{Tatoeba}}} \\
% \cline{2-9} 
% & \small{\textbf{fr-en}} & \small{\textbf{de-en}} & \small{\textbf{zh-en}} & \small{\textbf{de-en}} & \small{\textbf{jp-en}} & \small{\textbf{es-en}} & \small{\textbf{fr-en}} & \small{\textbf{hi-en}} \\
% \hline 
% \small{mE5-base} & \small{96.7} &\small{99} &\small{97.8} & \small{97.9}& \small{91}&\small{97.1}&\small{92.7} &\small{95.7}\\
% \small{mE5-par-model-A} & \small{97.4} &\small{99.3} &\small{98.4} & \small{98.5}& \small{90.9}&\small{97.4}&\small{93.1} &\small{95.5}\\
% \small{mE5-par-model-B} & \small{97} &\small{99} &\small{98.1} & \small{97.8}& \small{89.8}&\small{97}&\small{92.4} &\small{94}\\
% \hline 
% \end{tabular}
% \end{adjustbox}
% \caption{\label{table:bm-me5}Results for bitext mining for multilingual E5 model}
% \end{table}

\begin{table}[!htbp]
\centering
\begin{adjustbox}{max width=\textwidth}
\begin{tabular}{|p{0.25\linewidth}|>{\centering\arraybackslash}p{0.15\linewidth}|>{\centering\arraybackslash}p{0.15\linewidth}|>{\centering\arraybackslash}p{0.15\linewidth}|}
\hline 
\small{\textbf{lang pair}} & \small{\textbf{mE5-base}} & \small{\textbf{mE5-par-A}} & \small{\textbf{mE5-par-B}} \\
\hline
\multicolumn{4}{|c|}{\small{\textbf{BUCC}}} \\
\hline
\small{fr-en} & \small{96.7} & \small{97.4} & \small{97} \\
\small{de-en} & \small{99} & \small{99.3} & \small{99} \\
\small{zh-en} & \small{97.8} & \small{98.4} & \small{98} \\
\hline
\multicolumn{4}{|c|}{\textbf{Tatoeba}} \\
\hline
\small{de-en} & \small{97.9} & \small{98.5} & \small{97.8} \\
\small{jp-en} & \small{91} & \small{90.9} & \small{89.8} \\
\small{es-en} & \small{97.1} & \small{97.4} & \small{97} \\
\small{fr-en} & \small{92.7} & \small{93.1} & \small{92.4} \\
\small{hi-en} & \small{95.7} & \small{95.5} & \small{94} \\
\hline 
\end{tabular}
\end{adjustbox}
\caption{\label{table:bm-me5}Results for bitext mining for multilingual E5 model}
\end{table}

\subsection{Visualization}
To visualize the impact of alignment on cross-lingual similarity, we plotted histograms of cosine similarity scores for sentence pairs from the alignment development set before and after alignment. Specifically, for a given language pair (A, B), we randomly sampled 100 sentences in language A and paired them with their corresponding translations in language B to form 100 matched pairs. To create a contrastive set, we then shuffled the translations, forming 100 random pairs where sentences were no longer aligned.

We computed cosine similarity scores for both sets using XLMR-base before and after alignment and plotted histograms for different language pairs from our pool, as shown in Figure \ref{fig:histo-de-jp}, and more in appendix \ref{app:plots} . The results clearly show that post-alignment, cosine similarity scores for matched pairs become higher and well separated from those of random pairs, demonstrating improved cross-lingual alignment.
\begin{figure}[h]
  \centering
  \includegraphics[width=\linewidth]{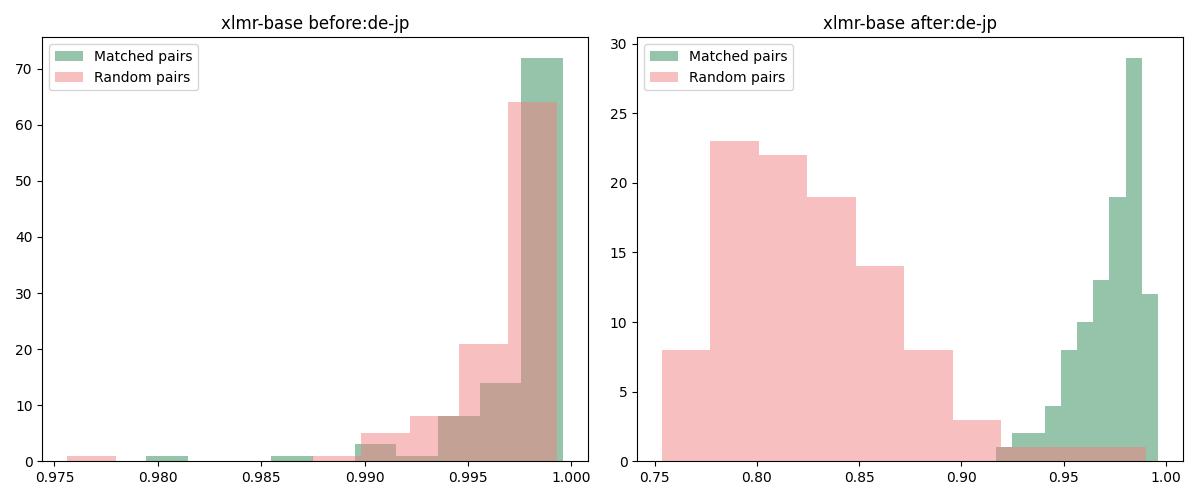} 
  \caption{Histograms of cosine similarity scores before and after alignment for German and Japanese sentences}
  \label{fig:histo-de-jp}
\end{figure}

%Add detail on the mixed eval set that was created for model selection:bfazili/evaluating-verifiability-in-generative-search-engines/verifiability_judgments/data_prep/mixup-for-model-selection.py

\section{Related Work}
\label{sec:related-work}

% Explicit Cross-Lingual Alignment
Early work on explicit cross-lingual alignment training was introduced by \citet{cao2020multilingualalignmentcontextualword}, which proposed the task of "contextual word retrieval" to optimize word correspondences across a bilingual parallel corpus using contextual word embeddings. \citet{wu-dredze-2020-explicit} extended this to multi-way parallel texts, but found limited gains when evaluated across diverse models and datasets.

% Word-Level and Dictionary-Based Alignment
Several approaches improve cross-lingual alignment at the word level. DICT-MLM~\cite{chaudhary2020dictmlmimprovedmultilingualpretraining} applies MLM while translating masked tokens using a bilingual dictionary. Align-MLM~\cite{tang2022alignmlmwordembeddingalignment} enhances word embedding alignment by encouraging similarity between dictionary pairs via an auxiliary loss. Other methods include dual-encoder masking~\cite{li-etal-2023-dual} and pre-alignment via code-switching~\cite{li-etal-2024-prealign}. Unlike these, our method retrofits alignment into a pretrained model using a small curated corpus, avoiding changes to the original (and expensive) pretraining process.

% Sentence-Level Multilingual Representations
Sentence-level cross-lingual alignment has also been widely studied. \citet{artetxe-schwenk-2019-massively} trained a single BiLSTM encoder with a shared BPE vocabulary and decoder. \citet{feng2022languageagnosticbertsentenceembedding} trained language-agnostic embeddings using a mix of MLM, TLM, and translation ranking with margin loss. Sentence-BERT (SBERT)~\cite{reimers2019sentencebertsentenceembeddingsusing} and its multilingual extension~\cite{reimers2020makingmonolingualsentenceembeddings} leverage NLI data and knowledge distillation. LASER and LaBSE are strong multilingual baselines but require large-scale parallel corpora—223M for LASER and 6B for LaBSE—and often struggle with semantically similar but non-identical sentence pairs.

\section{Conclusion}
Cross-lingual alignment ensures that semantically similar meanings across languages have closer representations. While previous work has focused on aligning bilingual text pairs, we show that training on multi-way parallel text—where all translations of a semantic instance are aligned together—is more effective. By leveraging translations from an MT model and allowing all languages to serve as anchors, our approach enhances cross-lingual alignment and improves downstream performance across multiple tasks in the MTEB benchmark. We also find notable improvements even for languages not included in the alignment training. This demonstrates the broader generalization impact of cross-lingual alignment with the proposed multi-way parallelism in the alignment dataset.
\section{Limitations}
\begin{enumerate}
    \item Our approach relies on a multi-way parallel corpus, synthesized by translating English source text into six target languages using the NLLB model~\cite{nllbteam2022languageleftbehindscaling}. However, we do not analyze the sensitivity of our training to translation quality. With consistently strong improvements across languages, we do not observe clear trends for lower-resource languages like Chinese, where translation quality may lag behind that of 
    languages closer to English, such as German or French.
    \item This study focuses on a limited set of six languages (in addition to English) as a preliminary investigation. Extending our approach to a massively multilingual setting remains an open direction for future work.
    \item Our alignment experiments are conducted on a relatively small dataset of $\sim{76K}$ rows of four-way parallel sentences in multiple languages. Scaling up the alignment corpus to larger sizes and adapting the training techniques to different kinds of models is left for future work.
    % potentially reaching performance levels comparable to state-of-the-art (SOTA) methods, 
    % remains unexplored—particularly given that our current model is significantly more lightweight and computationally efficient.
    \item The alignment technique presented here is designed for multilingual understanding tasks and may not directly benefit generative tasks, as suggested by \cite{li2023doeszeroshotcrosslingualgeneration}. Future research could explore evaluating this alignment on generative tasks and adapting the training strategy to generative models.
    % \item Although we took a representative subset of tasks from MTEB , evaluation on more tasks for MTEB or other benchmarks like XLNI could be included. 
\end{enumerate}

\bibliography{anthology,custom}
\appendix
\section{Model Training details}
\label{app:training_details}
For all models, no auxiliary layers or parameters were added while finetuning for alignment end-to-end. The loss was applied over the pooler output with 768 dimensions. Early stopping with patience of 10 over 20 epochs with batch size of 32 was implemented for XLMR-base and mBERT-base. For mE5, batch size of 128 was used over 5 epochs. We additionally boosted positive pairs with lower cosine similarity during contrastive learning by scaling respective logits with 
$\exp(-z_i.z_p)$. Temperature and regularisation scalar were carefully tuned for each of the models to select the best performing aligned version. Note that we maintain anchors in the contrastive loss from each language in both models described in Section \ref{sec:parallelism}. We conducted experiments using a NVIDIA A100-SXM4-80GB GPU with CUDA 12.4 and using PyTorch 2.6.

\section{More plots}
\label{app:plots}
\begin{figure}[h]
  \centering
  \includegraphics[width=\linewidth]{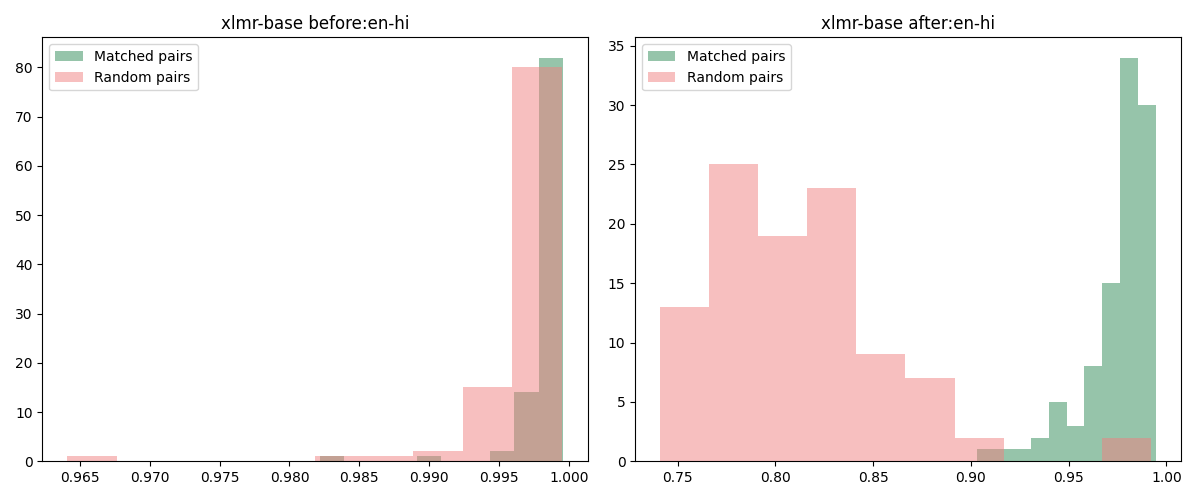} 
  \caption{Histograms of cosine similarity scores before and after alignment for English and Hindi sentences}
  \label{fig:histo-en-hi}
\end{figure}

\begin{figure}[h]
  \centering
  \includegraphics[width=\linewidth]{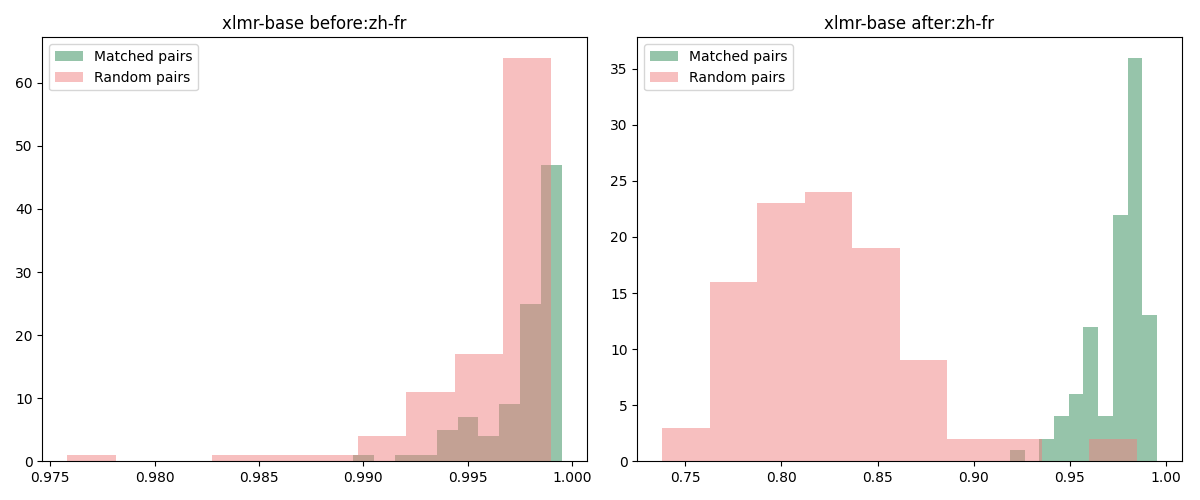} 
  \caption{Histograms of cosine similarity scores before and after alignment for Chinese and French sentences}
  \label{fig:histo-zh-fr}
\end{figure}

\begin{figure}[ht]
  \centering
  \includegraphics[width=\linewidth]{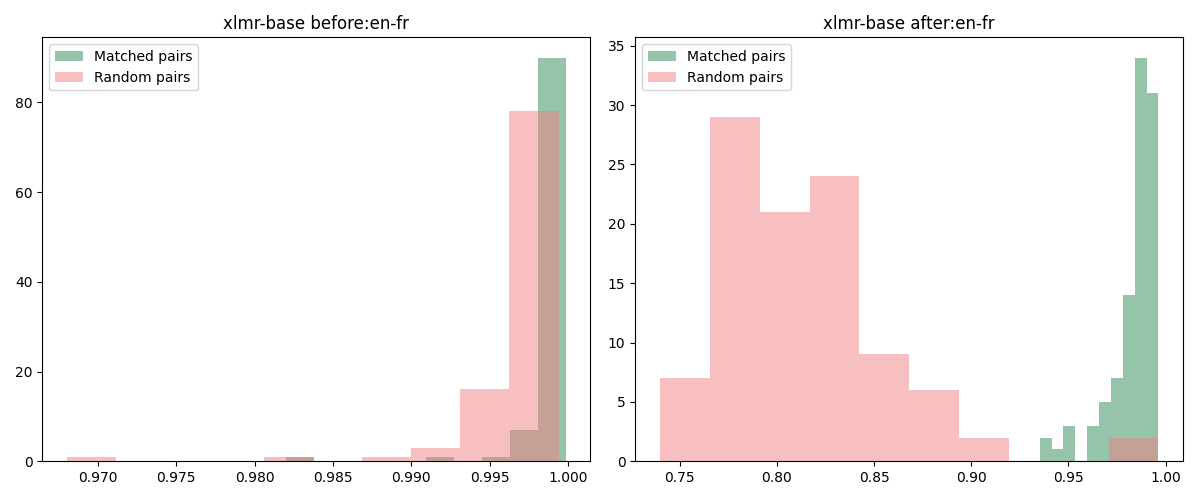} 
  \caption{Histograms of cosine similarity scores before and after alignment for English and French sentences}
  \label{fig:histo-en-fr}
\end{figure}

\begin{figure}[ht]
  \centering
  \includegraphics[width=\linewidth]{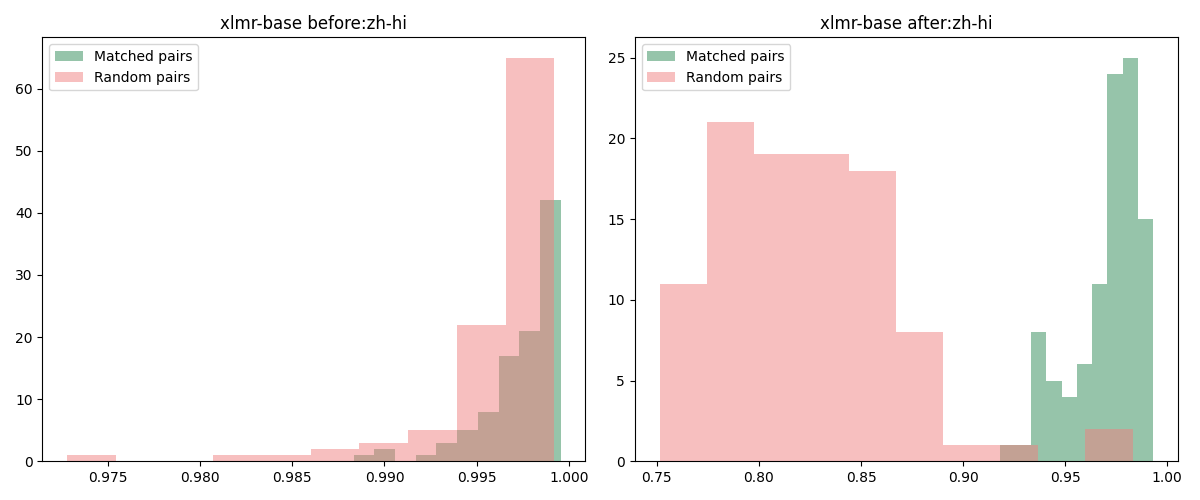} 
  \caption{Histograms of cosine similarity scores before and after alignment for Chinese and Hindi sentences}
  \label{fig:histo-hi-zh}
\end{figure}

Figures \ref{fig:histo-zh-fr}  shows how the range of similarity scores for matched Chinese and French sentences get well separated and higher post alignment from random pairs of sentences from the two languages. Similarly for English-Hindi in \ref{fig:histo-en-hi},English-French in Figure \ref{fig:histo-en-fr} and for Chinese-Hindi in Figure \ref{fig:histo-hi-zh}.


\end{document}